\documentclass[11pt]{article}
\usepackage{graphicx}
\usepackage{longtable,lscape}
\usepackage{amsfonts}
\usepackage{bbm}
\usepackage{float}
\usepackage{url}

\newcommand{\captionfonts}{\footnotesize}
\makeatletter  
\long\def\@makecaption#1#2{%
  \vskip\abovecaptionskip
  \sbox\@tempboxa{{\captionfonts #1: #2}}%
  \ifdim \wd\@tempboxa >\hsize
    {\captionfonts #1: #2\par}
  \else
    \hbox to\hsize{\hfil\box\@tempboxa\hfil}%
  \fi
  \vskip\belowcaptionskip}
\makeatother 
\begin{document}
\title{The Guppy Effect as Interference}
\author{Diederik Aerts$^1$, Jan Broekaert$^1$, Liane Gabora$^2$ and Tomas Veloz$^2$ \vspace{0.5 cm} \\ 
        \normalsize\itshape
        $^1$Center Leo Apostel for Interdisciplinary Studies \\
        \normalsize\itshape
        and Department of Mathematics, Brussels Free University \\ 
        \normalsize\itshape
         Krijgskundestraat 33, 1160 Brussels, Belgium \\
        \normalsize
        E-Mails: \url{diraerts@vub.ac.be,jbroekae@vub.ac.be} \\ \\
        \normalsize\itshape
        $^2$Department of Psychology and Mathematics, \\
        \normalsize\itshape
         University of British Columbia, Kelowna, British Columbia, Canada \\
        \normalsize
        E-Mails: \url{liane.gabora@ubc.ca, tomas.veloz@ubc.ca}
        }
\date{}
\maketitle
\begin{abstract}
\noindent
People use conjunctions and disjunctions of concepts in ways that violate the rules of classical 
logic, such as the law of compositionality. Specifically, they overextend conjunctions of concepts, a phenomenon referred to as the Guppy Effect. We build on previous efforts to develop a quantum model \cite{aertsgabora2005a,aerts2011,aerts2009a}, that explains the Guppy Effect in terms of interference. Using a well-studied data set with 16 exemplars that exhibit the Guppy Effect, we developed a 17-dimensional complex Hilbert space ${\cal H}$ that models the data and demonstrates the relationship between overextension and interference. We view the interference effect as, not a logical fallacy on the conjunction, but a signal that out of the two constituent concepts, a \emph{new} concept has emerged.
\end{abstract}


\section{The Guppy Effect -- Introduction\label{intro}}
A concrete formal understanding of how concepts combine is vital to significant progress in many fields including psychology, linguistics, and cognitive science. However, concepts have been resistant to mathematical description because people use conjunctions and disjunctions of concepts in ways that violate the rules of classical logic; i.e., concepts interact in ways that are non-compositional \cite{hampton1987}. This is true also with respect to properties (e.g., although people do not rate \emph{talks} as a characteristic property of {\it Pet} or {\it Bird}, they rate it as characteristic of {\it Pet Bird}) and exemplar typicalities (e.g., although people do not rate {\it Guppy} as a typical {\it Pet}, nor a typical {\it Fish}, they rate it as a highly typical {\it Pet Fish} \cite{oshersonsmith1981}). This has come to be known as the Pet Fish Problem, and the general phenomenon wherein the typicality of an exemplar for a conjunctively combined concept is greater than that for either of the constituent concepts has come to be called the Guppy Effect, although further investigation revealed that the Pet Fish Problem is not a particularly good example of the Guppy Effect, and that other concept combinations exhibit this effect more strongly \cite{stormsetal1998}. 

One can refer to the situation wherein people estimate the typicality of an exemplar of the concept combination as more extreme than it is for one of the constituent concepts in a conjunctive combination
as {\it overextension}. One can refer to the situation wherein people estimate the typicality of the exemplar for the concept conjunction
as higher than that of {\it both} constituent concepts as {\it double overextension}. We posit that overextension is not a violation of the classical logic of conjunction, but that it signals the emergence of a whole new concept. The aim of this paper is to model the Guppy Effect as an interference effect using a mathematical representation in a complex Hilbert space and the formalism of quantum theory to represent states and calculate probabilities. This builds on previous work that shows that Bell Inequalities are violated by concepts
\cite{aertsetal2000,gaboraaerts2002} and in particular by concept combinations that exhibit the Guppy Effect 
\cite{aertsgabora2005a,aerts2011,aerts2009a,aerts2007,aerts2009b}, and add to the investigation of other approaches using interference effects in cognition \cite{franco2009,francozuccon2010,lambertmogiliansky2009}.

Our approach is best explained with an example. Consider the data in Tab. 1. It is based on data obtained by asking participants to estimate how typical various exemplars are of the concepts {\it Furniture}, {\it Household Appliances}, and {\it Furniture and Household Appliances} \cite{hampton1988a}. \begin{table}[H]
\small
\begin{center}
\begin{tabular}{|clccccccc|}
\hline 
\multicolumn{2}{|l}{} & \multicolumn{1}{l}{$\mu(A)_k$} & \multicolumn{1}{l}{$\mu(B)_k$} & \multicolumn{1}{l}{$\mu(A\ {\rm and}\ B)_k$} & \multicolumn{1}{l}{${\mu(A)_k+\mu(B)_k \over 2}$} & \multicolumn{1}{l}{$\theta_k$} &\multicolumn{1}{l}{$\lambda_k$}&\multicolumn{1}{l|}{$\beta_k$} \\
\hline
\multicolumn{9}{|l|}{\it $A$=Furniture, $B$=Household Appliances} \\
\hline
1&{\it Filing Cabinet}&0.079&0.040&0.062&0.059&87.61 &-0.056 &-87.61\\
2&{\it Clothes Washer}&0.026&0,118&0.078&0.072&84.01 &0.055 &84.01\\
3&{\it Vacuum Cleaner}&0.017&0,118&0.051&0.068&112.21 &-0.042 &-112.21\\
4&{\it Hifi}&0.056&0.079&0.090&0.067&70.58 &0.063 &70.58\\
5&{\it Heated Waterbed}&0.089&0.050&0.082&0.070&79.28 &-0.066 &-79.28\\
6&{\it Sewing Chest}&0.075&0.058&0.061&0.067&94.74 &0.066 &94.74\\
7&{\it Floor Mat}&0.052&0.023&0.031&0.037&100.87&-0.034 &-100.87\\
8&{\it Coffee Table}&0,100&0.025&0.050&0.062&104.78&0.048 &104.78\\
9&{\it Piano}&0.084&0.020&0.043&0.052&101.67&0.040 &101.67\\
10&{\it Rug}&0.056&0.019&0.028&0.037&106.58&0.031 &106.58\\
11&{\it Painting}&0.057&0.014&0.021&0.035&120.16&-0.024 &-120.16\\
12&{\it Chair}&0.099&0.030&0.047&0.065&109.41&-0.052 &-109.41\\
13&{\it Fridge}&0.042&0,117&0.085&0.079&85.23&0.070 &85.23\\
14&{\it Desk Lamp}&0.066&0.079&0.085&0.072&79.85&-0.071 &-79.85\\
15&{\it Cooking Stove}&0.037&0,118&0.088&0.078&81.57&-0.066 &-81.57\\
16&{\it TV}&0.065&0.092&0.099&0.078&61.89&0.075 &61.89\\
\hline
\end{tabular}
\end{center}
\caption{Interference data for concepts {\it A=Furniture} and {\it B=Household Appliances}. The probability of a participant choosing exemplar $k$ as an example of {\it Furniture} 
or {\it Household Appliances} is given by $\mu(A)_k$ or ($\mu(B)_k$, respectively. The probability of a participant choosing a particular exemplar $k$ as an example of {\it Furniture and Household Appliances} is $\mu(A\ {\rm and}\ B)_k$. The classical probability would be ${\mu(A)_k+\mu(B)_k \over 2}$. The quantum phase angle $\theta_k$ 
introduces a quantum interference effect. Values are approximated to their third decimal, and angles to their second decimal.
}
\end{table}
\normalsize
\noindent
Although Hampton's original data was in the form of typicality estimates, for the quantum model that we built it is more appropriate for data to be in the form of `good examples'. 
Thus we calculated from Hampton's typicality data estimates for the following experimental situation. Participants are given the list of exemplars in Tab. 1 and asked to answer the following questions. {\it Question $A$} is `Choose one exemplar that you consider a good example of {\it Furniture}'. {\it Question $B$} is `Choose one exemplar that you consider a good example of {\it Household Appliances}'. Finally, {\it Question $A$ and $B$} is `Choose one exemplar that you consider a good example of {\it Furniture and Household Appliances}'.
Hence, concretely, the data in Tab. 1 were not collected by asking the three `good example'-questions but calculated from Hampton's data, derived from an experiment in which participants were asked to give typicality estimates. 
This transformation of Hampton's data retains the basic pattern of results because estimated typicality of an exemplar is strongly correlated with the frequency with which it is chosen as a good example \cite{hampton2007}.

\section{A Quantum model}
In this section we build a quantum model of the Guppy Effect by modeling Hampton's data in complex Hilbert space for the pair of concepts {\it Furniture} and {\it Household Appliances}, and their conjunction {\it Furniture and Household Appliances}. 
The way in which we calculated the `good example' data from Hampton's `typicality' data is by normalizing for each exemplar the typicality estimates of each participant giving rise to an estimate 
of the extent to which this exemplar constitutes a `good exemplar'. We then average on all the participants obtaining $\mu(A)_k$, $\mu(B)_k$ and $\mu(A\ {\rm and}\ B)_k$ (see Tab. 1). We interpret the resulting values as estimates
of the probability that exemplar $k$ is chosen as an answer for \emph{Questions} $A$, $B$, and `$A\ {\rm and}\ B$', respectively. Tab. 1 gives the probabilities of responses.
Hampton's original typicality data, which ranged between -3 and +3, were rescaled to a [0, 6] Likert scale to avoid negative values, and then afterwards normalized and averaged for each of the three concepts ($A$, $B$ and, `$A\ {\rm and}\ B$') (see Tab.1).

The `good example' measurement has 16 possible outcomes, namely each of the considered exemplars, and hence is represented in quantum theory by means of a self-adjoint operator with spectral decomposition $\{M_k\ \vert\ k=1,\ldots,16\}$ where each $M_k$ is an orthogonal projection of the Hilbert space ${\cal H}$ corresponding to exemplar $k$ from the list in Tab. 1. The concepts {\it Furniture} and {\it Household Appliances} are represented by orthogonal unit vectors $|A\rangle$ and $|B\rangle$ of the Hilbert space ${\cal H}$, and the combination {\it Furniture and Household Appliances} is represented by ${1 \over \sqrt{2}}(|A\rangle+|B\rangle)$, which is the normalized superposition of $|A\rangle$ and $|B\rangle$. 
It is by means of this superposition that the quantum framework can describe how a new concept `$A$ and $B$', emerges out of $A$ and $B$. In the following, the standard rules of quantum mechanics are applied to calculate the probabilities, $\mu(A)_k$, $\mu(B)_k$ and $\mu(A\ {\rm and}\ B)_k$
\begin{eqnarray}
&&\mu(A)_k=\langle A|M_k|A\rangle \quad \mu(B)_k=\langle B|M_k|B\rangle \\
 \label{interferencegeneral}
\mu(A\ {\rm and}\ B)_k&=&{1 \over 2} (\langle A \vert +\langle B \vert) M_k ( \vert A \rangle +\vert B \rangle)={1 \over 2}(\langle A|M_k|A\rangle+\langle B|M_k|B\rangle+\langle A|M_k|B\rangle+\langle B|M_k|A\rangle) \nonumber \\
&=&{1 \over 2}(\mu(A)_k+\mu(B)_k)+\Re\langle A|M_k|B\rangle \label{muAandB}
\end{eqnarray}
where $\Re\langle A|M_k|B\rangle$ is the interference term. Let us introduce $|e_k\rangle$ the unit vector on $M_k|A\rangle$ and $|f_k\rangle$ the unit vector on $M_k|B\rangle$, and put $\langle e_k|f_l\rangle= \delta_{kl} c_ke^{i\gamma_k}$. 
Then we have $|A\rangle=\sum_{k=1}^{16}a_ke^{i\alpha_k}|e_k\rangle$ and $|B\rangle=\sum_{k=1}^{16}b_ke^{i\beta_k}|f_k\rangle$, and with $\phi_k=\beta_k-\alpha_k+\gamma_k$, this gives 
\begin{eqnarray} 
\langle A|B\rangle&=&(\sum_{k=1}^{16}a_ke^{-i\alpha_k}\langle e_k|)(\sum_{l=1}^{16}b_le^{i\beta_l}|f_l\rangle)=\sum_{k=1}^{16}a_kb_kc_ke^{i(\beta_k-\alpha_k+\gamma_k)}=\sum_{k=1}^{16}a_kb_kc_ke^{i\phi_k} \label{ABequation} \\
\mu(A)_k&=& (\sum_{l=1}^{16}a_le^{-i\alpha_l}\langle e_l|)(a_ke^{i\alpha_k}|e_k\rangle)=a_k^2 \\
\mu(B)_k&=& (\sum_{l=1}^{16}b_le^{-i\beta_l}\langle f_l|)(b_ke^{i\beta_k}|f_k\rangle)=b_k^2 \\
\langle A|M_k|B\rangle&=& (\sum_{l=1}^{16}a_le^{-i\alpha_l}\langle e_l|)M_k|(\sum_{m=1}^{16}b_me^{i\beta_m}|f_m\rangle)=a_kb_ke^{i(\beta_k-\alpha_k)}\langle e_k|f_k\rangle=a_kb_kc_ke^{i\phi_k}
\end{eqnarray}
which, making use of (\ref{muAandB}), gives
\begin{equation} \label{muAandBequation}
\mu(A\ {\rm and}\ B)_k={1 \over 2}(\mu(A)_k+\mu(B)_k)+c_k\sqrt{\mu(A)_k\mu(B)_k}\cos\phi_k
\end{equation}
We choose $\phi_k$ such that
\begin{equation} \label{cosequation}
\cos\phi_k={2\mu(A\ {\rm and}\ B)_k-\mu(A)_k-\mu(B)_k \over 2c_k\sqrt{\mu(A)_k\mu(B)_k}}
\end{equation}
and hence (\ref{muAandBequation}) is satisfied. We now have to determine $c_k$ in such a way that $\langle A|B\rangle=0$. Note that from $\sum_{k=1}^{16}\mu(A\ {\rm and}\ B)_k=1$ and (\ref{muAandBequation}), and with the choice of $\cos\phi_k$ made in (\ref{cosequation}), it follows that \\ $\sum_{k=1}^{16}c_k\sqrt{\mu(A)_k\mu(B)_k}\cos\phi_k=0$. Taking into account (\ref{ABequation}), which gives $\langle A|B\rangle=\sum_{k=1}^{16}a_kb_kc_k(\cos\phi_k+i\sin\phi_k)$, and making use of $\sin\phi_k=\pm\sqrt{1-\cos^2\phi_k}$, we have
\begin{eqnarray}
&&\langle A|B\rangle=0
\Leftrightarrow
\sum_{k=1}^{16}c_k\sqrt{\mu(A)_k\mu(B)_k}(\cos\phi_k+i\sin\phi_k)=0 \\
&&\Leftrightarrow\sum_{k=1}^{16}c_k\sqrt{\mu(A)_k\mu(B)_k}\sin\phi_k=0 \\ \label{conditionequation}
&&\Leftrightarrow\sum_{k=1}^{16}\pm\sqrt{c_k^2\mu(A)_k\mu(B)_k-(\mu(A\ {\rm and}\ B)_k-{\mu(A)_k+\mu(B)_k \over 2})^2}=0
\end{eqnarray}
We introduce the following quantities
\begin{equation} \label{lambdak}
\lambda_k=\pm\sqrt{\mu(A)_k\mu(B)_k-\left(\mu(A\ {\rm and}\ B)_k-{\mu(A)_k+\mu(B)_k \over 2}\right)^2}
\end{equation}
and choose $m$ the index for which $|\lambda_m|$ is the biggest of the $|\lambda_k|$'s. Then we take $c_k=1$ for $k\not=m$. 
We now explain the algorithm used to choose a plus or minus sign for $\lambda_k$ as defined in (\ref{lambdak}), with the aim of being able to determine $c_m$ such that (\ref{conditionequation}) is satisfied. 

We start by choosing a plus sign for $\lambda_m$. Then we choose a minus sign in (\ref{lambdak}) for the $\lambda_k$ for which $|\lambda_k|$ is the second biggest; let us call the index of this term $m_2$. This means that $0\le\lambda_m+\lambda_{m_2}$. For the $\lambda_k$ for which $|\lambda_k|$ is the third biggest -- let us call the index of this term $m_3$ -- we choose a minus sign 
if $0\le\lambda_m+\lambda_{m_2} - \vert \lambda_{m_3}\vert$, and otherwise we choose a plus sign, and in the present case we have $0 > \lambda_m+\lambda_{m_2}- \vert \lambda_{m_3}\vert$. 
We continue this way of choosing, always considering the next biggest $|\lambda_k|$, and hence arrive at a global choice of signs for all of the $\lambda_k$, such that $0\le\lambda_m+\sum_{k\not=m}\lambda_k$. Then we determine $c_m$ such that (\ref{conditionequation}) is satisfied, or more specifically such that
\begin{equation} \label{cmequation}
c_m=\sqrt{{(-\sum_{k\not=m}\lambda_k)^2+(\mu(A\ {\rm and}\ B)_m-{\mu(A)_m+\mu(B)_m \over 2})^2 \over \mu(A)_m\mu(B)_m}}
\end{equation}
We choose the sign for $\phi_k$ as defined in (\ref{cosequation}) equal to the sign of $\lambda_k$. The result of the specific solution thus constructed is that we can take $M_k({\cal H})$ to be rays of dimension 1 for $k\not=m$, and $M_m({\cal H})$ to be a plane. This means that we can make our solution still more explicit. Indeed, we take ${\cal H}={\mathbb{C}}^{17}$, the canonical 17-dimensional complex Hilbert space, and make the following choices
\begin{eqnarray} 
&|A\rangle= \left(\sqrt{\mu(A)_1},\ldots,\sqrt{\mu(A)_m},\ldots,\sqrt{\mu(A)_{16}},0 \right) \label{vectorA} \\
&|B\rangle= \left(e^{i\beta_1}\sqrt{\mu(B)_1},\cdots,c_me^{i\beta_m}\sqrt{\mu(B)_m},\cdots, \right. \nonumber \\ 
& \left. e^{i\beta_{16}}\sqrt{\mu(B)_{16}},\sqrt{\mu(B)_m(1-c_m^2)}\right) \label{vectorB} \\ 
&\beta_m=\arccos \left({2\mu(A\ {\rm and}\ B)_m-\mu(A)_m-\mu(B)_m \over 2c_m\sqrt{\mu(A)_m\mu(B)_m}}\right) \label{anglebetan} \\
&\beta_k=\pm\arccos \left({2\mu(A\ {\rm and}\ B)_k-\mu(A)_k-\mu(B)_k \over 2\sqrt{\mu(A)_k\mu(B)_k}}\right) \label{anglebetak}
\end{eqnarray}
where the plus or minus sign in (\ref{anglebetak}) is chosen following the algorithm introduced for choosing the plus and minus sign for $\lambda_k$ in (\ref{lambdak}). Let us construct this quantum model for the data in Tab. 1. The exemplar that gives the biggest value of $|\lambda_k|$ is {\it TV}, and hence we choose a plus sign and get $\lambda_{16}=0.0745$. The exemplar that gives the second biggest value of $\lambda_k$ is {\it Desk Lamp}, and hence we choose a minus sign, and get $\lambda_{14}=-0.0710$. Next comes {\it Fridge} having $|\lambda_{13}|=0.0698$, and since $\lambda_{16}+\lambda_{14}<0$, we choose a plus sign for $\lambda_{13}$. We determine in a recursive way the signs for the remaining exemplars. Tab. 1 gives the values of $\lambda_k$ calculated following this algorithm. From (\ref{cmequation}) it follows that $c_{16}=0.564$.

Making use of (\ref{vectorA}), (\ref{vectorB}), (\ref{anglebetak}) and (\ref{anglebetan}), and the values of the angles given in Tab. 1, we put forward the following explicit representation of the vectors $|A\rangle$ and $|B\rangle$ in ${\mathbb{C}}^{17}$ representing concepts {\it Furniture} and {\it Household appliances}. 
\begin{eqnarray}
&|A\rangle=(0.280, 0.161, 0.131, 0.236, 0.299, 0.274, 
0.229, 0.316, 0.289, 0.236, 0.238, \nonumber \\
& 0.315, 0.205, 0.257, 0.193, 0.255, 0) \\
&|B\rangle=(0.200 e^{-i 87.61^{\circ}}, 0.343e^{i 84.01^{\circ}}, 0.343e^{-i 112.20^{\circ}}, 
0.281e^{i 70.58^{\circ}}, 0.225e^{-i 79.28^{\circ}}, \nonumber \\
& 0.242e^{i 94.73^{\circ}}, 0.151e^{-i 100.87^{\circ}}, 0.157e^{i 104.78^{\circ}}, 0.140e^{i 101.67^{\circ}}, 
0.137e^{i 106.58^{\circ}}, \nonumber \\ 
& 0.119e^{-i 120.16^{\circ}}, 0.174e^{-i 109.41^{\circ}}, 0.342e^{i 85.23^{\circ}}, 0.280e^{-i 79.85^{\circ}}, \nonumber \\
& 0.344e^{-i 81.57^{\circ}}, 0.171e^{i 61.89^{\circ}}, 0.250).
\end{eqnarray}
This proves it is possible to make a quantum model of the \cite{hampton1988a} data such that the values of $\mu(A\ {\rm and}\ B)_k$ are determined from the values of $\mu(A)_k$ and $\mu(B)_k$ as a consequence of quantum interference effects. For each exemplar $k$, the value of $\theta_k$ in Tab. 1 gives the quantum interference phase. 

\section{Visualization of Interference Probabilities}
A previous paper provided a quantum representation of the concepts {\it Fruits} and {\it Vegetables} and their disjunction {\it Fruits or Vegetables}, and gave a way to graphically represent possible quantum interference patterns that result when concepts combine \cite{aerts2009b}.
Here we follow this procedure to generate a graphical representation for the concepts {\it Furniture}, {\it Household Appliances}, and their conjunction {\it Furniture and Household Appliances}. Each concept is represented by complex valued wave functions of two real variables $\psi_A(x,y)$, $\psi_B(x,y)$ and $\psi_{A {\rm and} B}(x,y)$. We choose $\psi_A(x,y)$ and $\psi_B(x,y)$ such that the square of the absolute value of both wave functions is a Gaussian in two dimensions, which is always possible since we only have to fit 16 values, namely those of $|\psi_A|^2$ and $|\psi_B|^2$ for each of the exemplars of Tab. 1. These Gaussians are graphically represented in Figs. 1 a) and 1 b), and the exemplars of Tab. 1 are located in spots such that the Gaussian distributions $|\psi_A(x,y)|^2$ and $|\psi_B(x,y)|^2$ properly model the probabilities $\mu(A)_k$ and $\mu(B)_k$ in Tab. 1 for each of the exemplars.
\noindent 
\begin{figure}
\centerline {\includegraphics[height=19cm,width=12.5cm]{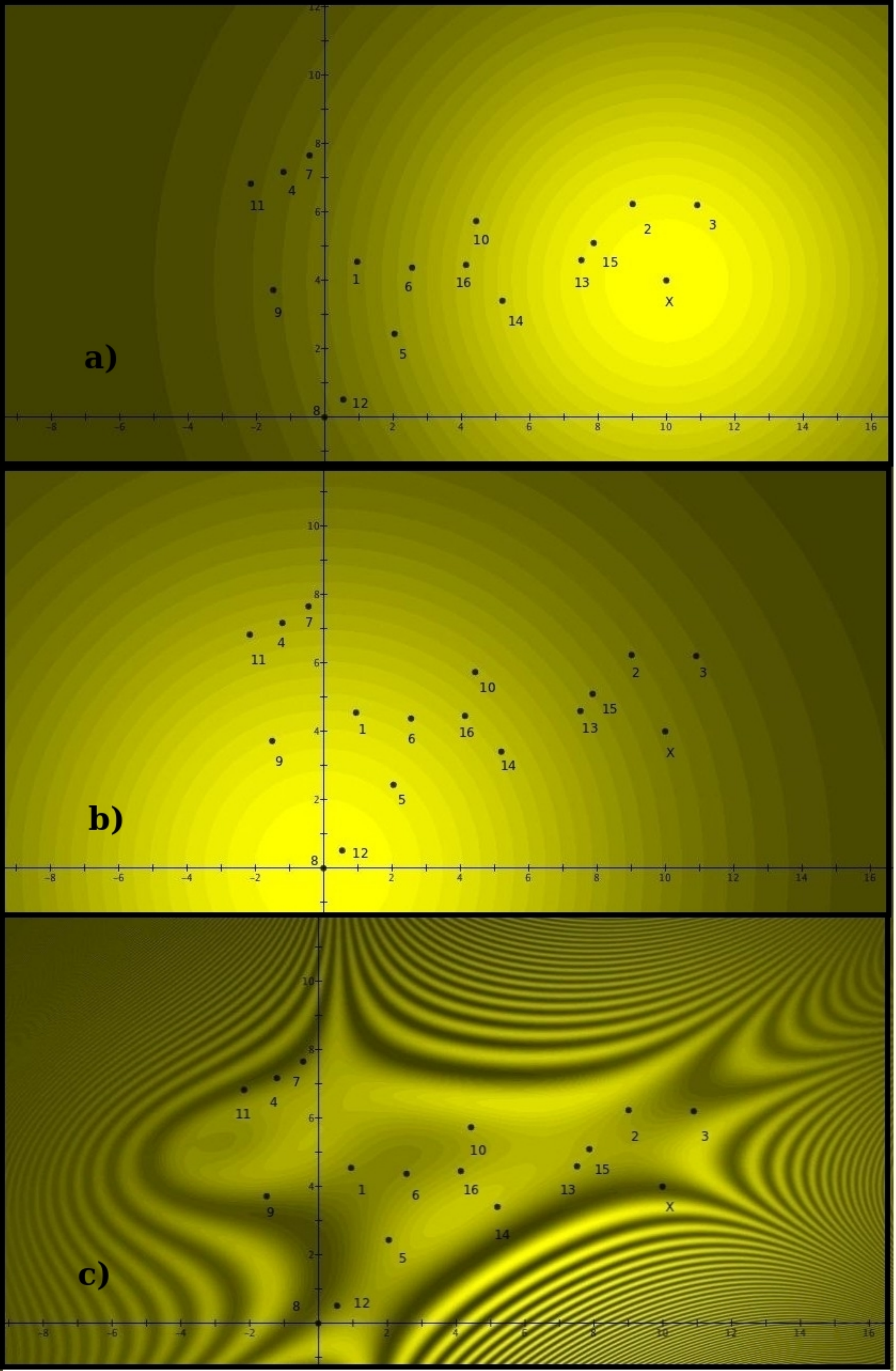}}
\caption{A representation of our quantum model of {\it Furniture}, {\it Household Appliance} and {\it Furniture and Household Appliance} by a double slit interference situation. 
The brightness of the light source in a
region corresponds to the probability that an exemplar in this region is chosen as a `good example' of the concept {\it Furniture} in figure a), {\it Household Appliance} in figure b), and {\it Furniture and Household Appliance} in figure c). Numbers indicate the exemplars as numbered in Table 1.
}
\end{figure}
For example, for {\it Furniture} (Fig. 1 a)), {\it Coffee Table} is located in the centre of the Gaussian because it was most frequently chosen in response to {\it Question A}. {\it Chair} was the second most frequently chosen, hence it is closest to the top of the Gaussian. Note that in Fig. 1 b) there is one point labelled by $X$, which is the maximum of the Gaussian representing $\mu(B)$. We preferred not to locate the highest value of typicality by the maximum of the Gaussian, because doing so did not lead to an easy fit of both Gaussians. For {\it Household Appliances}, represented in Fig. 1 b), $X$ is located in the maximum of the Gaussian, and since {\it Clothes Washer}
 and {\it Vacuum Cleaner} are the most frequently chosen (with exactly the same frequency) they are located closest to $X$ at an equal distance radius. {\it Cooking Stove} was the third most frequently chosen, then {\it Fridge} and so on, with {\it Painting} as the least chosen `good examples' of {\it Household Appliances}. Metaphorically, we could regard the graphical representations of Figs. 2 a), 2 b) as the projections of a light source shining through two holes such that a screen captures it and the holes make the intensity follow a Gaussian distribution when projected on the screen.
The centre of the first hole, corresponding to {\it Furniture}, is located where exemplar {\it Coffee Table} is at point $(0, 0)$, indicated by $8$ in both figures. The centre of the second hole, corresponding to {\it Household Appliances}, is located where point {\it X} is at (10,4), indicated by 17 in both figures. In Fig. 1 c) the data for {\it Furniture and Household Appliances} are graphically represented. This is not `just' a normalized sum of the two Gaussians of Figs. 2 a) and b), since it is the probability distribution corresponding to ${1 \over \sqrt{2}}(\psi_A(x,y)+\psi_B(x,y))$, which is the normalized superposition of the wave functions in Figs. 2 a) and b). The numbers are placed at the locations of the different exemplars, according to the labels of Tab. 1, with respect to the probability distribution ${1 \over 2}|\psi_A(x,y)+\psi_B(x,y)|^2={1 \over 2}(|\psi_A(x,y)|^2+|\psi_B(x,y)|^2)+|\psi_A(x,y)\psi_B(x,y)|\cos\theta(x,y)$, where $|\psi_A(x,y)\psi_B(x,y)|\cos\theta(x,y)$ is the interference term and $\theta(x,y)$ the quantum phase difference at $(x,y)$. The values of $\theta(x,y)$ are given in Tab. 1 for the locations of the different exemplars.
The interference pattern in Fig. 1 c) is very similar to well-known interference patterns of light passing through an elastic material under stress. In our case, it is the interference pattern corresponding to {\it Furniture and Household Appliances}. Bearing in mind the analogy with the light source and holes for Figs. 1 a) and b), in Fig. 1 c) we can see the interference pattern produced when both holes are open. (For the mathematical details -- the exact form of the wave functions and the calculation of possible interference patterns -- and other examples of conceptual interference, see \cite{aerts2009b}.)

\section{Interpretation of Interference in Cognitive Space}
If we consider equations (\ref{interferencegeneral}) and (\ref{muAandBequation}), the fundamental interference equations used in this quantum model, we see that $\mu(A\ {\rm and}\ B)$ becomes equal to the average ${1 \over 2}(\mu(A) + \mu(B))$ in case of 
no interference, i.e., if the interference terms are zero. Thus the description of the conjunction as a no interference situation does not coincide with what is obtained using the minimum rule from fuzzy 
set theory. Note that in the double slit situation, a 
classical particle passing through with both slits 
open gives rise to a probability distribution on the screen which is 
equal to ${1 \over 2}(\mu(A) + \mu(B))$, i.e. the average of the 
probabilities with only one of the two slits open. Hence, in both the interference quantum model and its double slit representation, the average plays the role 
of the classical default, not the minimum, as one would expect to be the case if the conjunction were modeled using fuzzy set theory.

This aspect of our model needs further explanation. First, as in an earlier interference based model of disjunction  \cite{aerts2009b}, for the conjunction, the average is the 
classical default, not the maximum, as would follow from a fuzzy set theory model. Second, if we consider Hampton's data,
the average $1/2(\mu(A)+\mu(B))$ is effectively closer to the frequency of the combined concept $\mu(A\ {\rm and}\ B)$ than the fuzzy set minimum value. More concretely, on average, the probability for the combined concept differs 0.011 from the classical average, but 0.026 from the fuzzy set minimum measure (Fig.-Tab. 2). Also, calculation of the correlation between the probability for the combined concept and the average and the minimum, yields 0.899 and 0.795 respectively, which indicates that experimentally the average is a better estimate than the minimum.

The findings that (1) the average is the classical default in our quantum model, and in the double slit representation of it, and (2) the average is also a better experimental approximation than the minimum, indicate that the connective `and' in a conjunction of concepts does not play the role that we imagine it to play intuitively and from our experience with logic. 
Similarly, the connective `or' in a disjunction of concepts does not play the role we imagine it to play \cite{aerts2009b}. A similar phenomenon was identified for Hampton's data on membership weights of exemplars with respect to conjunctive and disjunctive combinations of pairs of concepts. This was resolved by showing that the state space is a Fock space with two sectors, the first sector describing this `non logical and interference role' of conjunction and disjunction, with indeed the average as classical default, and a second sector describing the logical role of conjunction and disjunction, with minimum and maximum as classical defaults in the case of conjunction and respectively disjunction, and quantum entanglement as a quantum effect \cite{aerts2009a}. We believe that this is also the state of affairs here, and that we have only described the `first sector Fock space' part in the present article, hence the interference part, with the average as classical default, and a role of conjunction that is not the one of logic. Since the present model describes the interference part in the first sector of Fock space, but not the entanglement part in the second sector of Fock space, it can be seen as complementary to an entanglement quantum model that was worked out for the Pet-Fish concept combination in a tensor product Hilbert space \cite{aertsgabora2005a}.
\begin{figure}[H]
\begin{center}
\includegraphics[height=7cm,width=17.5cm]{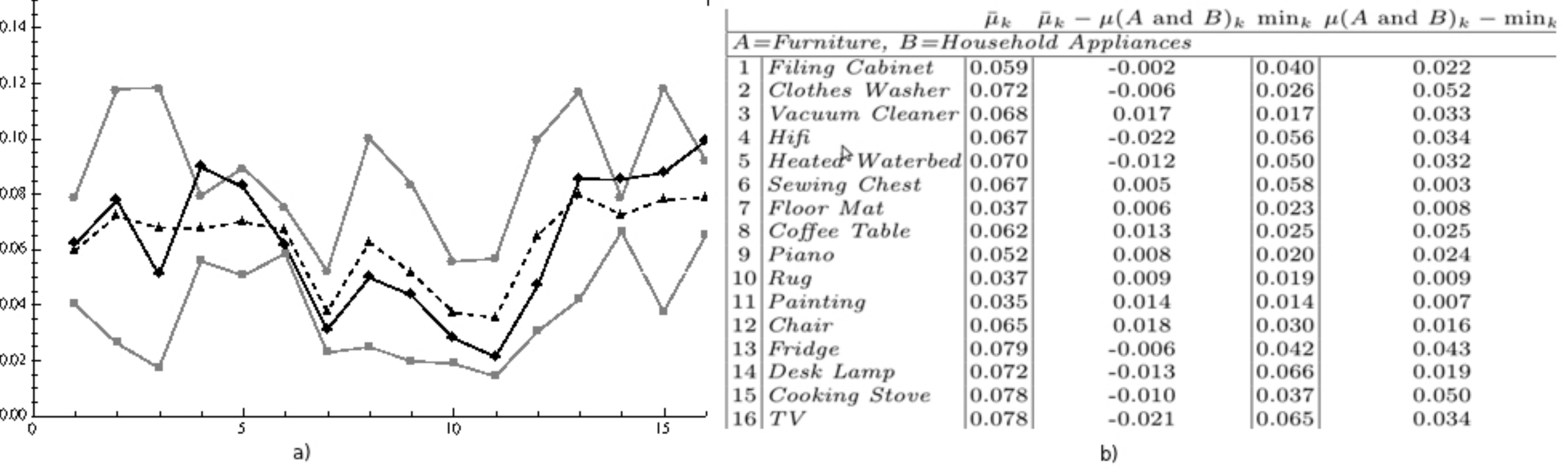}
\caption{a) Estimations of concept combination probabilities. The horizontal axis corresponds to the exemplar label denoted by $k$ in Tab.1 and the vertical axis measures estimated probability. The two grey curves represent the minimum and maximum for each $k$ of the probabilities $\mu(A)_k$ and $\mu(B)_k$. The black curve represents the probability $\mu(A\ {\rm and}\ B)_k$ obtained from the data, and the dashed curve represents the average between $\mu(A)_k$ and $\mu(B)_k$. b) Comparison between the concept conjunction probability, and the classical average and minimum probabilities of the concepts {\it A} and {\it B}. $\bar\mu_k=\frac{\mu(A)_k+\mu(B)_k}{2}$ is the the classical average probability (third column) and ${\rm min}_k= {\rm min}\{ \mu(A)_k ,\mu(B)_k\}$ is the minimum probability (fifth column). The fourth (sixth) column shows the deviation of the average (the minimum) with respect to the concept combination probability. The probability $\mu(A\ {\rm and}\ B)_k$ deviates $0.011$ from the average $\bar\mu_k$, but 0.026 from the minimum.}
\end{center}
\end{figure}
\vspace{-0.4cm}
\noindent
An more intuitive way of looking at this is that when it comes to first sector of Fock space effects, hence interference effects, participants mainly consider 
`{\it Furniture and Household Appliances}' in its root combination `{\it Furniture--Household Appliances}', without taking into account the `and' as a logical connective. The `and' merely introduces an extra context on this root combination, which, for example, will be different from the extra context introduced by the `or' on the root combination.

At first sight it may seem that our interference quantum model does not incorporate order effects, which are known to exist experimentally. More concretely, experiments on the combination `$A$ and $B$' will often lead to different data than experiments on the combination `$B$ and $A$'. However, order effect can be modeled without problems in our interference approach, because in the first sector of Fock space, although `$A$ and $B$' and `$B$ and $A$' are described by the same superposition state, the phase of this state is different, leading to different interference angles, and hence different values for the collapse probabilities. This is how the first sector of a Fock space interference model copes in a natural way with order effects. 

The double slit representation also helps clarify aspects of the situation
and thus provides new insight into concept combination. 
The role of the two slits is played by {\it Furniture} and {\it Household Appliances}, and the role of the specific positions on the detection screen where the interference pattern is formed are played by the measuring locations for the exemplars. We can see clearly that the mind is not working with these concepts in a classical manner. If this were the case, each individual would simply substitute the combined concept {\it Furniture and Household Appliances} by one of the two constituent concepts -- in a manner similar to how the classical particle passes through one of the two slits. This would result in a perfect average, and hence no interference. 
However, on many occasions individual judgements of typicality for the conjunction deviate from the average, in a manner similar to how the statistical average of typicality deviates from the average. 
This means that interference is operating, similar to the interference pattern observed in quantum mechanics even with single quantum `particles' in a double slit set-up \cite{donatimissirolipozzi1973}. Physicists introduced the term `self-interference' to indicate this behavior. The above suggests that the individual ponders each of the constituents of a combined concept and this process takes place `in superposition' when referring to the individual constituents of the combination. This is the expression of the emergence of a new concept for this combination.
Other aspects of the origin of conceptual interference effects, and their implications for cognition and creativity, are analyzed and discussed elsewhere \cite{aerts2011,aerts2009a,aerts2007,aertsdhooghe2009,Velozetal2011}.

Let us finish this section by returning to the issue of overextension of the conjunction. Since from our analysis it follows that the average, a first order classical default (namely the default of the first sector of Fock space), is stronger than the minimum, the classical default (of the second sector of Fock space), the notion of `overextension' no longer covers correctly the `deviation from classicality'. However, an interesting relation with interference can be found. 
Overextension takes place when
\begin{equation}
\mu(A\ {\rm and}\ B)_k - {\rm min } \{\mu(A)_k, \mu( B)_k\} > 0
\end{equation}
which is equivalent to 
\begin{equation}
{\rm max } \{\mu(A)_k, \mu( B)_k\} > \frac{ \mu(A)_k + \mu( B)_k}{2} - \Re\langle A|M_k|B\rangle
\end{equation}
Overextension occurs when the average modulated by the interference term cannot equal the largest of the constituent typicalities.
This is consistent with the contention that `Conjunctions tend to be overextended to include exemplars that are good members of one class, but are marginal to the other'~\cite{hampton1988a}. Only for double overextension is interference necessary. On the other hand, situations where one concept in the conjunction is  very atypical, while the other is highly typical, a situation traditionally considered unproblematic, could require a large interference deviation from the classical average.

\section{Conclusions}
We presented a quantum model that demonstrates how the Guppy Effect can be modeled as interference. A data set for two concepts and their conjunction -- with an ontology of 16 exemplars -- was modeled in a 17-dimensional Hilbert space ${\cal H}$. The non-compositionality of the conjunction of concepts was identified by its close convergence to the classical average of probabilities, while the quantum interference appears as a modulation to fit the effect of the logical connectives.
Our core finding is that this effect produces a quantifiable deviation from classical analyses, signalling the emergence of a new concept. 
One implication is that in some situations, particularly when new content emerges, cognitive processes cannot be described using classical logic.

\end{document}